\documentclass[10pt,twocolumn,letterpaper]{article}

\usepackage{cvpr}
\usepackage{times}
\usepackage{epsfig}
\usepackage{graphicx}
\usepackage{amsmath}
\usepackage{amssymb}
\usepackage{subfiles}
\usepackage{algorithm}
\usepackage{algorithmic}
\usepackage{multirow}

\usepackage{array}
\usepackage{dsfont}
\usepackage{subfigure}
\usepackage[abs]{overpic}

\usepackage{soul}

\usepackage[pagebackref=true,breaklinks=true,letterpaper=true,colorlinks,bookmarks=false]{hyperref}

 \cvprfinalcopy % *** Uncomment this line for the final submission

\def\cvprPaperID{2451} % *** Enter the CVPR Paper ID here

\ifcvprfinal\fi
\begin{document}

%%%%%%%%% TITLE
\title{Intrinsic Image Transformation via Scale Space Decomposition}

\author{Lechao Cheng$^{1}$ \hspace{20mm}  Chengyi Zhang$^{1}$ \hspace{22mm} Zicheng Liao$^{1, 2}$\vspace{2mm}\\
College of Computer Science, Zhejiang University$^{1}$\vspace{0mm}\\
Alibaba - Zhejiang University Joint Institute of Frontier Technologies$^{2}$
}

\maketitle
%\thispagestyle{empty}

%%%%%%%%% ABSTRACT
\begin{abstract}
We introduce a new network structure for decomposing an image into its intrinsic albedo and shading. We treat this as an image-to-image transformation problem and explore the scale space of the input and output. By expanding the output images (albedo and shading) into their Laplacian pyramid components, we develop a multi-channel network structure that learns the image-to-image transformation function in successive frequency bands in parallel, within each channel is a fully convolutional neural network with skip connections. This network structure is general and extensible, and has demonstrated excellent performance on the intrinsic image decomposition problem.  We evaluate the network on two benchmark datasets: the MPI-Sintel dataset and the MIT Intrinsic Images dataset. Both quantitative and qualitative results show our model delivers a clear progression over state-of-the-art. %We also show qualitative results that are visually close to the ground-truth.
\end{abstract}

\section{Introduction}\label{sec:intro}
%To disentangle the intrinsic and periphery information from an image of a scene -- such as the reflectance, normals and illumination -- is an fundamental problem in computer vision, as it unfolds the mechanism of the biological vision system.

There has been an emerging trend in representation learning that learns to disentangle from an image latent codes accounting for the various dimensions of the input, e.g., illumination, pose, and attributes~\cite{DBLP:BaoCWLH17,DBLP:ShuYHSSS17,disentangled:cvpr17}. %It remains a core problem in the vision community to disentangle intrinsic components from an image, e.g. the material and the shading
%While much progress has been seen,
Yet one of the preliminary forms of this problem -- to decompose an image into its intrinsic \emph{albedo} and \emph{shading} -- has drawn less attention. Solutions to the intrinsic image decomposition problem would enable material editing, provide cues for depth estimation, and provide a computational explanation to the long standing lightness constancy problem in perception. However, even with exciting progress (e.g.~\cite{Chen:iccv13,DBLP:KimPSL16}), this problem still remains a challenging task for continuing effort.
%early day Retinex model to the most recent feed-forward neural networks~\cite{land1971lightness,Barron:2012A,Chen:iccv13,DBLP:KimPSL16} still produce results to be improved.

%Difficult because of under-determinedness and lack of data.
%Under-determinedness; piecewise smooth assumption not universally true (holds only for lambertian reflectance and restrained geometric property, e.g. a Manhattan world); the results are images (albedo and shading) that encapsulate complicated transformation from material, geometry and optical interaction of the input image. To model this relation, we need data to provide sufficient supervision and a model that can embrace the complexity of this problem.
Part of the difficulty arises from the under-determinedness of this problem. Based on prior knowledge of albedo and shading, the Retinex algorithm constrains the decomposition into a thresholding problem in the gradient domain. This model is practical, but would fail to handle complex material or geometry that has sharp edges or casts shadows under strong point sources. Another part of the difficulty lies in the complexity of the forward image generation process -- a process that transforms scene material, geometry and illumination into a 2D image via the dynamics of optical interactions and projection. Intrinsic image decomposition is partly trying to \emph{invert} this process.

\begin{figure}
  \centering
  \begin{overpic}[width=1.0\linewidth,clip,trim=0 850 1450 0]{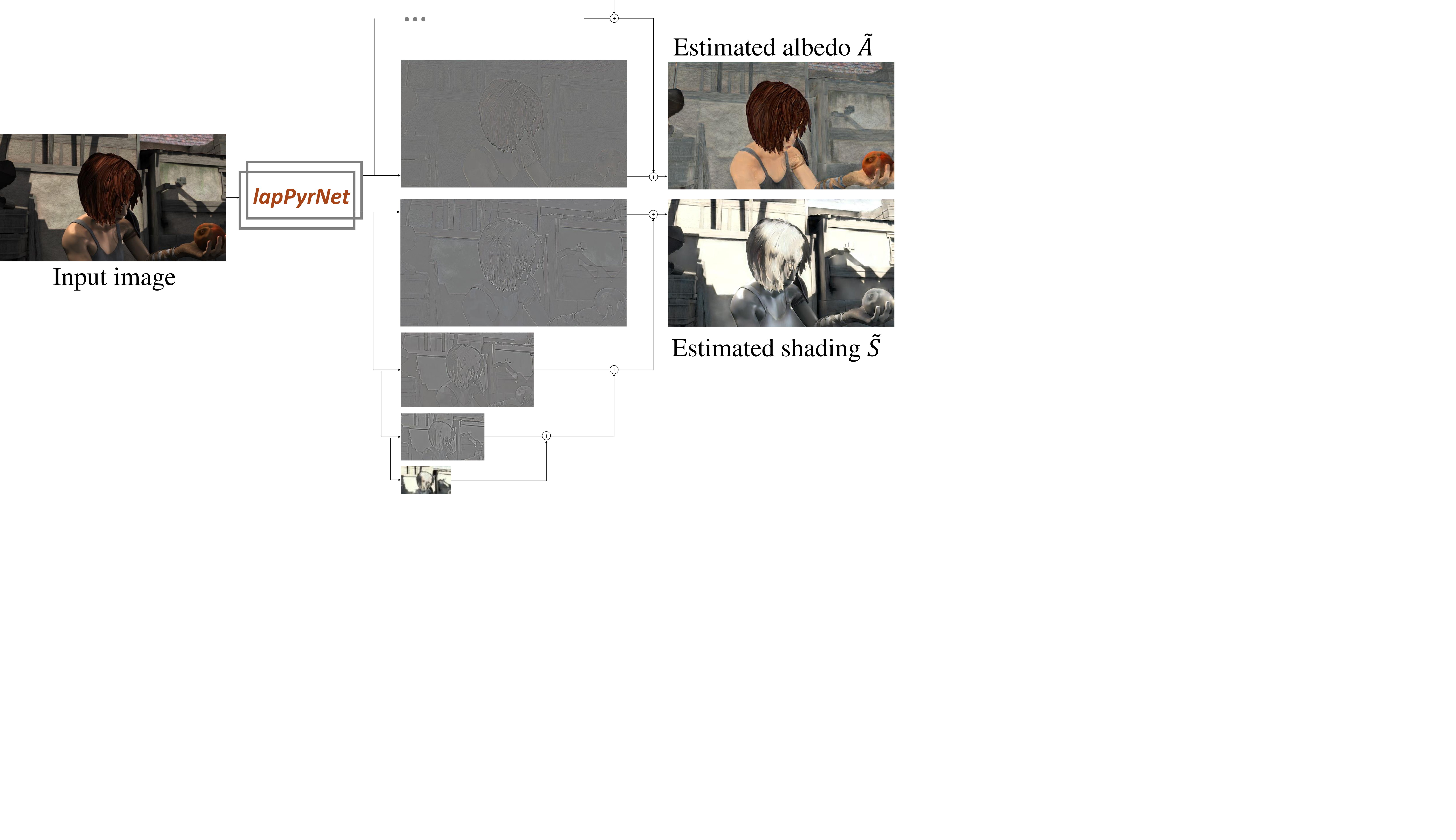}\end{overpic}\\
  \caption {Given an input image, our \emph{lapPyrNet} jointly produces Laplacian pyramid components that collapse into the target albedo and shading images in high quality. Our network features by a multi-channel architecture that treats intrinsic image decomposition as image-to-image transformation in separate frequency bands in the scale space.}\label{fig:illustration}\vspace{-4mm}
\end{figure}

%Main idea: formulate as an image-to-image transformation problem (that of the mapping from one image to output image(s)), exploiting LoD decomposition principle (expand the mapping to \emph{functions} operating in a successive frequency band, because an image has a natural scale space decomposition); data augmentation to boostrap existing training data.

In this work, we treat the intrinsic image decomposition process in an image-to-image transformation framework, using a deep neural network as a function approximator to learn the mapping relations. %Unlike previous networks (e.g.~\cite{narihira2015direct,lettry2016darn}), our work explores the scale space of the network input and output such that we expand the network into a parallel set of sub-band transformations. Another distinctive feature from previous work is a scheme of data augmentation that fights against the scarcity of labeled data of this task and allows us to incorporate unlimited unlabeled data for sampling of the image manifold.
While models of similar ideas have been proposed (e.g.~\cite{narihira2015direct,lettry2016darn}), %our model explores the scale space of the network input and output and expands the approximator pipeline into a parallel set of sub-band transformations. %This result in a multi-branch network that produces a level-of-detail decomposition of the output images.
our model explores the \emph{scale space} of the network input and output, and considers to simply the transformation as a whole by horizontally expanding the functor approximation pipeline into a parallel set of sub-band transformations. %Another desideratum of our model is a data augmentation scheme that fights against the scarcity of labeled data and allows us to incorporate unlimited unlabeled data for the sampling of the image manifold in training.

%we exploit the fact that both the network input and output are images and that images have a natural scale space decomposition, so that we expand the mapping to a series of functions operating in a successive series of non-overlapping frequency bands, reusing the Gaussian and Laplacian style pyramidal decomposition structure with learnable up/down samplers.

%Another desideratum is to fight against the scarcity of labeled data for network training. We take inspiration from \emph{breeder learning}~\cite{nair2008analysis} that uses a preliminary network to generate predictions from unlabeled images, and a \emph{synthesis} procedure to perturb and generate new data with exact ground truth labels for further training. This data augmentation scheme is applicable to other network training that learns to invert a generative process. %allows us to incorporate limitless amount of data for sufficient sampling of the image manifold.

%{\bf Contribution.} (1) formulate intrinsic decomposition as a joint image-to-image \emph{transformation} task; (2) exploiting the pyramidal structure of images, designing a network to produces LoD components of the output separately; (3) data-augmentation to fight with label scarcity; (4) state-of-the-art performance on benchmark dataset, i.e. the MPI-sintel dataset.

The contribution of this work is in developing a scale-space decomposition network for intrinsic image generation. %which in turn is applicable for other image-to-image transformation problems.
We do this by resuing the classical Gaussian and Laplacian pyramid structure with learnable down/up samplers. The result is a multi-branch network that produces a level-of-detail decomposition of the output albedo and shading; each decomposition component is predicted by one sub-network, which is aggregated together to match the target (Figure~\ref{fig:illustration}). We propose novel loss functions that respect the distinctive properties of albedo and shading for edge preservation and smoothness, respectively.
%In the data augmentation episode,
We further implement a data augmentation scheme to fight against the scarcity of labeled data -- that is, we take inspiration from \emph{breeder learning}~\cite{nair2008analysis}, and use a preliminarily trained network to generate predictions from unlabeled images, and a \emph{synthesis} procedure to perturb and generate new data with exact ground truth labels for iterative model refinement. This data augmentation scheme is applicable to other network training that learns to invert a generative process.

%\noindentv
%{\bf Limitations.} Our result is still...

We have evaluated our model on the MPI-Sintel dataset~\cite{butler2012naturalistic} and the MIT intrinsic image dataset~\cite{Grosse:2009}. Experimental results demonstrate the effectiveness of the proposed model and our network engineering components. Our final model achieves state-of-the-art performance with a significant margin over previous methods in a variety of evaluation metrics.

%%%%%%%%%%%%%%%%%%%%%%%%%%%%%%%%%%%%%%%%%%%%%%%%%%%%%%
%%%%%%%%%%%%%%%%%%%%%%%%%%%%%%%%%%%%%%%%%%%%%%%%%%%%%%
%%%%%%%%%%%%%%%%%%%%%%%%%%%%%%%%%%%%%%%%%%%%%%%%%%%%%%
\section{Related work}\label{sec:related}
\noindent
{\bf Intrinsic images: }
%The problem of intrinsic images is formulated as to recover the
A series of solutions have been seen since Barrow and Tenenbaum first propose this problem~\cite{tenenbaum2000global}, for example, the Retinex method~\cite{land1971lightness,grosse09intrinsic}, learning based method using local texture and color cues~\cite{Tappen:2006:pami}, and joint optimization using data-driven priors~\cite{Barron:2012A}. With the advent of deep neural networks, solution to this problem has shifted to a pure data-driven, end-to-end training with various forms of feed forward convolutional neural networks. Direct Intrinsics~\cite{narihira2015direct} is a successful early example of this type, using a (back then seemingly bold) multi-layer CNN architecture to transform an image directly into shading and albedo. Successive models include the work of Kim et al.~\cite{DBLP:KimPSL16} that predicts depth and the other intrinsic components together with a joint convolutional network that has shared intermediate layers and a joint CRF loss, and the DARN~\cite{lettry2016darn} network that incorporates a discriminator network and the adversarial training scheme to enhance of the performance of a ``generator'' network that produces the decomposition results.
%and the work of Fan et al.~\cite{DBLP:FanWHC17} learns an edge preserving image smoothing network with novel functional components and supervision channels
%Decomposition and disentangling.
%Traditional methods: Retinex assumption (rule-based)~\cite{grosse09intrinsic,land1971lightness};
%classification based on local texture and color cues~\cite{Tappen:2006:pami};
%joint optimization (data-driven prior + optimization)~\cite{Barron:2012A,Chen:iccv13};
%Deep networks (data-driven + image-to-image transformation)~\cite{narihira2015direct,lettry2016darn,DBLP:KimPSL16,DBLP:FanWHC17}.

\noindent
{\bf Image scale space and pyramid structures:}
The investigation of image scale space is no less old-fashioned than that of the intrinsic image decomposition in vision.
The studies of Koenderink~\cite{koenderink1984} in the 1980's reveals a diffusion process that ``explicitly defines the deep structure of (an) image'' that relates to the DOG structure revealed in even earlier studies~\cite{Marr1980}. Around the same time, Burt and Adelson proposed the Laplacian pyramid structure that decomposes an image into a hierarchical Level-Of-Detail (LOD) representation using successive Gaussian filtering and the DOG operator~\cite{BurtAdelson:1983}. Scale space decomposition also widely exists in other fields of study, such as 3D graphics (e.g.~\cite{Guskov:1999}) and numerical computing (e.g.~\cite{multigrid:wesseling04}).
%In computer graphics, interestingly, an anisotropic analog of the Laplacian pyramid -- the multiresolution ``detail vector'' representation  -- is developed for signal processing on meshes~\cite{Guskov:1999}.

%It may be of higher interest to see the use of hierarchical structures
Deep convolutional networks provide a natural hierarchical feature pyramid for multi-scale information processing.
The feature pyramid network (FPN) makes predictions from multi-level feature maps for object detection with top-down communication~\cite{lin2016feature}.
Pinheiro et al.~\cite{pinheiro2015learning} propose a two-way hierarchical feature aggregation network for object segmentation.
The work of Ghiasi et al.~\cite{ghiasi2016laplacian} produces segmentation score maps with spatia-semantic trade-offs from different network layers, and aggregates them into a final segmentation map by pyramid collapsing.
%Pyramid network for super resolution~\cite{lai2017deep}
The work of Lai et al.~\cite{lai2017deep} utilizes a similarly deeply stacked network and feature maps to generate image detail map of multi-scales for image super-resolution. Notably, all of the above work utilizes hierarchical features from a CNN network for multi-scale processing.
%Laplacian GAN~\cite{denton2015deep}
In generative modeling, a Laplacian pyramid inspired GAN network is proposed by Denton et al.~\cite{denton2015deep} that learns generative modules in a Laplacian pyramid structure for image generation.

%Scale space in image structure analysis~\cite{koenderink1984}
%Laplacian pyramid for image encoding~\cite{burt1983laplacian}.
%In graphics, LoD representation for mesh structure: base mesh plus a set of detail vectors;
%Multigrid method in numerical computation~\cite{multigrid:wesseling04}
%Inception network~\cite{lin2013network,szegedy2015going}

%\noindent
%{\bf Pyramidal structure in deep networks:} Feature pyramid networks~\cite{lin2016feature}

\noindent
{\bf Image-to-image transformation: }
There is a variety of vision tasks that can be formulated as image-to-image transformation problem. Intrinsic image decomposition is one such example. Isola et al.~\cite{isola2016image} recently introduced an image-to-image translation network for several other tasks, including image colorization, sketch-to-image, and image-to-map generation. In this work, Isola et al. model the image-to-image transformation as a conditional generative process and use an adversarial loss for network training. %Our network is largely tangential to the model in Isola et al.~\cite{isola2016image}.

Note that a set of other vision tasks, such as dense pixel labeling (e.g. object segmentation~\cite{IslamCVPR17}), depth estimation from single image~\cite{XuCVPR17depth}, and the recent label-to-image synthesis network (\cite{ChenK17aa}, also in~\cite{isola2016image}) can also be framed as the image-to-image transformation problem, that is, to map pixels to pixels. Instead of hand engineering the mapping process for each task individually, we engineered a generic, extensible network architecture that is tangential to the work of Isola et al.~\cite{isola2016image} and features in exploiting the dimension of scale-space decomposition for the form of input/output transformation of this problem.

%Intrinsic decomposition can be considered one of the image transformation problems where the output is a set of intrinsic components of the input image. Various network architectures have been seen for the more general of image-to-image transform problem~\cite{isola2016image}, including tasks like image colorization, image-to-map, and sketch-to-object generation.
%This transformation network architecture is generic, e.g. predict pixels from pixels. Application like depth estimation, dense pixel labeling (e.g. segmentation~\cite{ghiasi2016laplacian}), or the recent segmentation-to-image synthesis network~\cite{ChenK17aa}

%Our network utilizes the scale space decomposition property of the image-to-image transformation problem, so it is generically applicable. In the next section, we introduce the network architecture design process and its implementation details.

%A related problem is generative models for plausibly looking image synthesis, there the input is a vector $z$ from a latent space (i.e., GAN~\cite{goodfellow2014generative} or VAE~\cite{kingma2013auto}). Laplacian pyramid model for generative models~\cite{denton2015deep}.

%%%%%%%%%%%%%%%%%%%%%%%%%%%%%%%%%%%%%%%%%%%%%%%%%%%%%%
%%%%%%%%%%%%%%%%%%%%%%%%%%%%%%%%%%%%%%%%%%%%%%%%%%%%%%
%%%%%%%%%%%%%%%%%%%%%%%%%%%%%%%%%%%%%%%%%%%%%%%%%%%%%%
\section{Method}
Let us first consider the transformation of an input image $I$ to an output image $A$ as a complex, highly nonlinear, and pixel-wise nonlocal mapping function $I \rightarrow f(I)$. It has been well demonstrated that deep convolutional neural networks are a general and practical parametrization and optimization framework for a variety of such mapping relations (from image classification to image-to-language translation). Now, let us consider how to adapt the network architecture to the \emph{image-to-image} transformation problem, in which the input and output are both images that have a natural Level-Of-Detail (LOD) pyramid structure, and that the mapping function linking the input to the output may also have a multi-channel decomposition based on the pyramid hierarchy.
In the next section (\ref{sec:reformation}) we are going to describe our model reformation process from a ResNet architecture that exploits this property to our final multi-channel hierarchical network architecture.

We write the Gaussian pyramid of an image $I$ as $[I_0, I_1,..., I_K]$, where $I_0 = I$ and $K$ is the total number of layers. We denote the $k$-th Laplacian pyramid layer by $\mathcal{L}_{k}(I) = I_k - u(I_{k+1})$ where $u$ is the up-sample operator. By definition, the Laplacian pyramid expansion of the image is $I = [\mathcal{L}_0(I), \mathcal{L}_1(I),..., \mathcal{L}_{k-1}(I), I_K]$, where $\mathcal{L}_0(I)$ is the detail layer of the original resolution and $I_K$ is the lowest resolution layer defined in the Gaussian pyramid.

\begin{figure}
  \centering
  \begin{overpic}[width=1.0\linewidth,clip,trim=15 0 142 0]{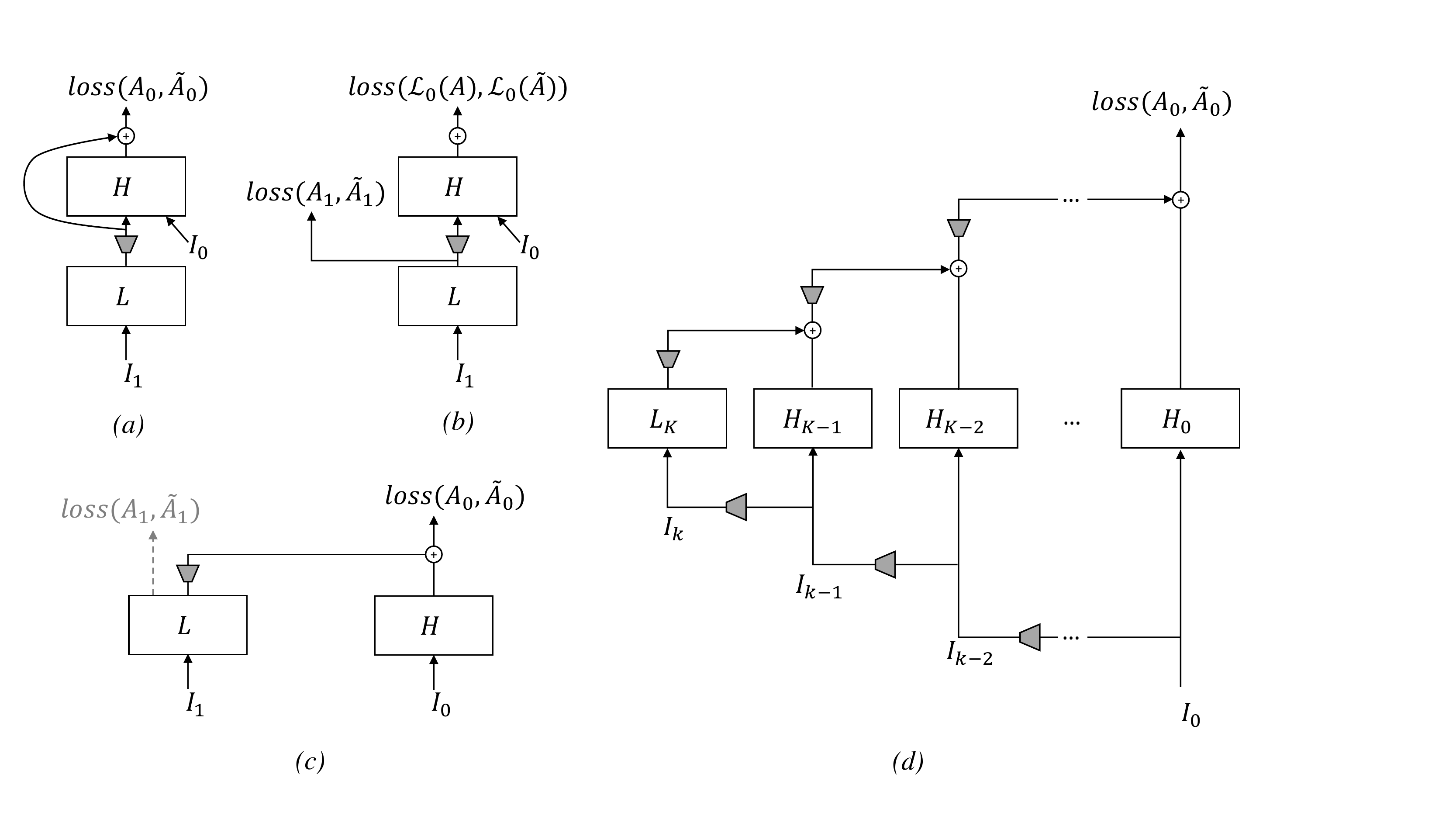}\end{overpic}\\
  \caption {Network architecture reformation (see section~\ref{sec:reformation}).}\label{fig:reformation}%\vspace{-3mm}
\end{figure}
\subsection{Network Architecture and Reformation}~\label{sec:reformation}
First, let us use a simplified network of two blocks ($L$ and $H$) to model the mapping $I \rightarrow f(I)$: $L$ for the mapping of the low frequency band, and $H$ handles the mapping in the high frequency band and whatever residuals that are left out by $L$. With the skip connection and summation of the output of $L$ to the output of $H$, this network (Figure~\ref{fig:reformation}-a) is an instantiation of the ResNet architecture~\cite{he2016deep}. %It is straightforward to unfold this architecture for multi-residual blocks, compared to which the ResNet architecture can be seen as with a more flexible skip connection scheme.

Next, by applying Laplacian pyramid expansion on the output, we can split the loss for (a) into two components: the output of $L$ is  restrained to fit the low-frequency Gaussian component, and that of $H$ to fit the Laplacian detail component separately (Figure~\ref{fig:reformation}-b). This reformed network is equivalent to (a) but with tighter constraints.

A critical transition is from (b) to (c) -- as it turns out to be possible to re-wire the two stacked blocks into two parallel branches, by connecting the output of $L$ to that of $H$ with summation, and adjusting the loss on $H$ accordingly. The resulted network structure (c) is equivalent to (b) -- they represent two equivalent forms of the Laplacian decomposition equation, i.e., by moving the residual component from lhs to rhs and change the sign. The loss of $L$ in (c) remains the same as a regularizer and our experiments find it is optional and is a barrier for numerical performance. %it turns out to be optional and a barrier for numerical performance.
The network structure (c) is the building block for our final extended model.

The final extended model is illustrated in Figure~\ref{fig:reformation}-d, for which we introduce multiple sub-network blocks $H_0,H_1,...H_{K-1}$ for the high frequency bands and one subnetwork block $L_K$ for the low frequency, in analogy to the Laplacian pyramid decomposition structure: the inputs to the network blocks are down-sampled in cascade, and outputs of the network blocks are up-sampled and aggregated from left to right to form the target output. All of the parameters of the down-sample and up-sample operators (the gray-shaded trapezoids in Figure~\ref{fig:reformation}) are learned in network. All of the network blocks share the same architectural topology, which we refer as ``residual blocks'' and describe in detail in section~\ref{sec:residual_block}. %In our intrinsic decomposition network, there is a pair of such networks, one for

\iffalse
\noindent
{\bf Discussion:} Note that our final network (d) in a non-obvious derivation from a ResNet counterpart, with a significant distinction that the sub-network blocks are not stacked up in sequence but wired \emph{in parallel}. The fact that this flattened parallel network structure relates to a deeply stacked network with skip connection may be worth of further investigation, for which we have not reached any conclusion. Besides, this parallel structure resembles that of a multi-branch network~\cite{xie2016aggregated}, but here each branch is a rather complex network module (see section~\ref{sec:residual_block}) instead of a few light-weighted convolutional filters as in \cite{xie2016aggregated}. Therefore, our network instantiation can be seen as a mixture of the designing principles of the multi-branch network~\cite{xie2016aggregated}, the Inception network~\cite{szegedy2015going}, and the Network-in-Network architecture~\cite{lin2013network}.
\fi

\subsection{Residual Block}\label{sec:residual_block}
\begin{figure}
  \centering
  \begin{overpic}[width=1.0\linewidth,clip]{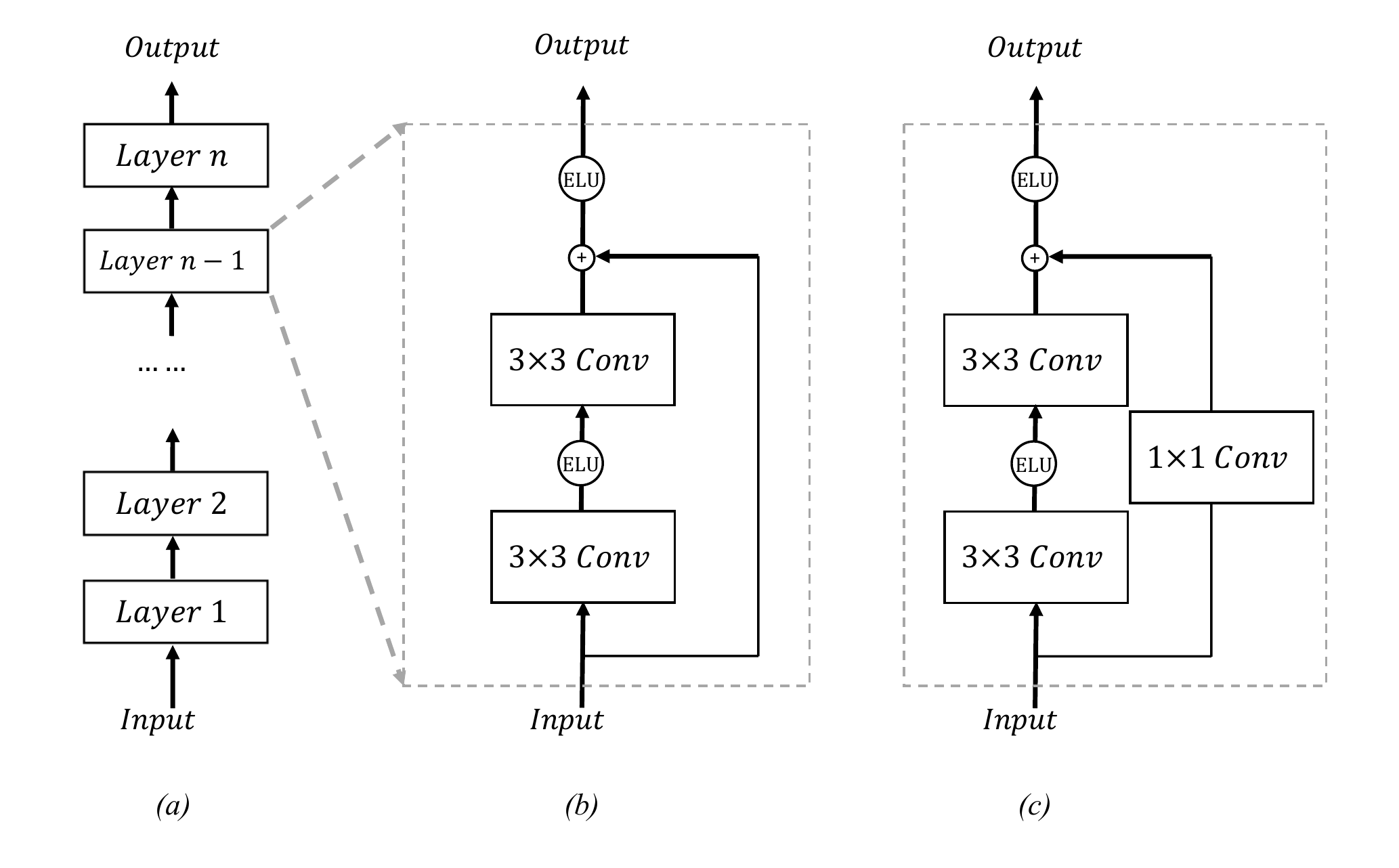}\end{overpic}\\
  \caption {Illustration of our Residual Block}\label{fig:residualblock}%\vspace{-3mm}
\end{figure}

The residual blocks are end-to-end convolutional subnetworks that share the same topology, and transform the input in different scales to the corresponding Laplacian pyramid components. Each residual block consists of 6 sequentially concatenated Conv(3x3)-ELU-Conv(3x3)-ELU sub-structures (Figure~\ref{fig:residualblock} (a-b)). Because we are predicting per-pixel value from an input image, no fully connected layers are used. We adopt the skip connection scheme that is popular in recent researches  (e.g. ~\cite{he2016deep}, ~\cite{lin2016feature}), including some variant of the DenseNet architecture by Huang et al.~\cite{huang2016densely}. Specifically, in each sub-structure, the output of the last Conv is element-wise accumulated with a skip connection, and the result is the input to the last ELU unit. The intermediate layers have 32 feature channels and output is a 3-channel image or residual image. A 1x1 Conv is added to the skip connection path of the first and last layer for dimension expansion/reduction to match the output of the residual path (Figure~\ref{fig:residualblock} (c)). %Our experimental result shows the skip connections yield performance convincingly better than classical ConvNets.

Instead of ReLU and Batch Normalization, we use Exponential Linear Units (ELU) as our activation function~\cite{Djork2016Fast}, because ELU can generate negative activation value when $x<0$ and has zero-mean activations, both of which improve the robustness to noise and convergence in training when our network becomes deeper. Besides, we removed the BN layer because it can be partially replaced by ELU which is 2x faster and more memory efficient.

\subsection{Loss Function}

%Being inspired by the design of loss function in previous work (e.g. ~\cite{DirectIntrinsic:2015}, ~\cite{Chen:iccv13}), we design a combined loss function, which could guarantee the semantic similarity, spatial resemblance and the smoothness of the images that our network predicts. We formally express our combined loss function to be optimized as:
The loss function is defined as follows:
\begin{align}
\mathcal{L} = \lambda_d\mathcal{L}_{data} + \lambda_p\mathcal{L}_{percep} + \lambda_t\mathcal{L}_{tv}
\end{align}
which contains a \emph{Data} loss, a feature-based \emph{Perceptual} loss and a \emph{Total Variation} loss as regularization. The hyper parameters are empirically set as: $\lambda_d$ = 1.0; $\lambda_p$ =0.5; $\lambda_t = 10^{-4}$.

\noindent{\bf Data loss:} %The Data Loss could reflect the pixel-level similarity between our prediction and the ground-truth. While the pyramid network architecture usually blurred some unimportant details in the final prediction, we could not apply classical error like mean square loss (MSE) directly. Instead, we consider applying the loss using \textit{cross bilateral filtering} as following:
The data loss defines pixel level similarity between the predicted image and the ground-truth. Instead of using the pixel-wise MSE, we employ the following \textit{joint bilateral filtering} (also known as \textit{cross bilateral filtering}\cite{eisemann2004flash,petschnigg2004digital}) loss combined with the constraint that the multiplication of the predicted albedo and shading should match the input:\\
\begin{align}
\color{red}
&\mathcal{L}_{data} = \sum\limits_{\mathcal{C} \in \{A,S\}}\frac{1}{N_p} \sum\limits_{p\in \mathcal{C}}\vert\vert{\mathcal{J}_p} - C_p\vert\vert_2^2 +\vert\vert \widetilde{A}\!*\!\widetilde{S}\!-\!{I}\vert\vert_2^2\\
&\mathcal{J}_p = \frac{1}{\mathcal{W}_p}\sum\limits_{q\in \mathcal{N}(p)}G_{\sigma_s}(\vert\vert p-q\vert\vert)G_{\sigma_r}(\vert \mathcal{C}_p - \mathcal{C}_q \vert)\widetilde{\mathcal{C}}_p\\
&\mathcal{W}_p= \sum\limits_{q\in \mathcal{N}(p)}G_{\sigma_s}(\vert\vert p-q\vert\vert)G_{\sigma_r}(\vert{\mathcal{C}}_p - \mathcal{C}_q \vert)
\end{align}

%\begin{align}
%&\mathcal{L}_{data}^{\mathcal{C}} = \frac{1}{N_p} \sum\limits_{p\in \mathcal{C}}\frac{1}{\mathcal{W}_p}\sum\limits_{q\in \mathcal{N}(p)}{F(C,\widetilde{C}, p, q)}\\
%&F(C,\widetilde{C}, p, q) = G_{\sigma_s}(\vert\vert p-q\vert\vert)G_{\sigma_r}(\vert \widetilde{\mathcal{C}}_p - \mathcal{C}_q \vert)\vert\vert\widetilde{\mathcal{C}}_p - \mathcal{C}_q\vert\vert \\
%&\mathcal{W}_p= \sum\limits_{q\in \mathcal{N}(p)}G_{\sigma_s}(\vert\vert p-q\vert\vert)G_{\sigma_r}(\vert\widetilde{\mathcal{C}}_p - \mathcal{C}_q \vert)
%\end{align}
%where $\mathcal{C} \in \{A,S\}$.
The cross bilateral filtering loss ensures smoothness of the output albedo and shading, and preserves sharp edges for albedo and strong cast shadows in shading (e.g. Figure~\ref{fig:illustration}). In contrary, the alternative MSE loss tends to produce blurry edges across boundaries in the output, which is also seen in \cite{narihira2015direct} and \cite{lettry2016darn} (see Figure~\ref{fig:comparison}). Here $\sigma_s$ = 1.0, $\sigma_r$ uses the adaptive bilateral filtering mechanism, $G_{\sigma_s}$ and $G_{\sigma_r}$ are the spatial and range Gaussian kernels, both with neighborhood size 5x5.

%From a perspective of graphics, an ideal prediction by our network could produce a pair of albedo and shading image, which could be recombined to the original image, by applying element-wise multiplication with a Retinex constraint, We would like to embody the importance of this relation, so we append an term in the final data loss as following:

%{\color{blue} \st{Further, we add the constraint that the multiplication of the predicted albedo and shading should match the input. So we train the albedo network and shading network \emph{jointly}, using data loss with the following three terms:}}
%\begin{align}
%    \mathcal{L}_{data} = \mathcal{L}_{data}^{A} + \mathcal{L}_{data}^{S} + \vert\vert \widetilde{A}\!*\!\widetilde{S}\!-\!{I}\vert\vert_2^2
%\end{align}

\noindent{\bf Perceptual loss:}
%High-level semantic information should be preserved in the transformation process, so a perceptual loss is widely used for protecting the stability of the semantic structure of the images(e.g. ~\cite{disentangled:cvpr17}). We make use of the standard VGG-19~\cite{Simonyan2014Very}, to extract high-level semantic information by detection the activation of neurons.We formally express the feature loss as following:
High-level semantic structures should be preserved in the transformation process as well, so a CNN-feature based perceptual loss~\cite{johnson2016perceptual,dosovitskiy2016generating} is used. We make use of the standard VGG-19~\cite{Simonyan2014Very} network to extract semantic information from neuron activations. Our perceptual loss is defined as follows:
\begin{align}
\mathcal{L}_{feat} = \sum\limits_{\mathcal{C} \in \{A,S\}}\sum\limits_{l} \frac{1}{F_lH_lW_l}\vert\vert \Phi_l(\widetilde{\mathcal{C}}) - \Phi_l(\mathcal{C}) \vert\vert_2^2
\end{align}
where $\Phi_l(\mathcal{C})$ is the network activations of $C$ at the $l$-th layer that have size $F_l\times H_l\times W_l$, and $l = relu1\_2, relu2\_2, relu3\_4$ and $relu4\_4$ are the VGG-19 network layers before pooling.

\noindent{\bf Total Variation loss:}
%In order to avoiding unimportant noises in predicted images, we using a regularization term to ensure the smoothness of them.In addition, a regularization term could also stabilize the training procedure of the network.Inspired by Rudin et al.~\cite{Rudin1992Nonlinear}, we would like to choose a \textit{total variation loss} as following:
Lastly, we use a total variation term to impose smoothness of the output results.
\begin{align}
\mathcal{L}_{tv} =  \sum\limits_{\mathcal{C} \in \{A,S\}}\sum\limits_{i,j}\vert\widetilde{\mathcal{C}}_{i+i,j}-\widetilde{\mathcal{C}}_{i,j}\vert + \vert\widetilde{\mathcal{C}}_{i,j+1}-\widetilde{\mathcal{C}}_{i,j}\vert
\end{align}
where $i$ and $j$ are image row and column indices.

Our final model is trained with the above loss on the output of $H_0$ combined with all outputs from lower level branches (Figure~\ref{fig:reformation}-(d)). This constrains all network channels simultaneously and gradients can back-propagate and dispatch more flexibly. Another training scheme, as we mentioned in section~\ref{sec:reformation}, is to train the network from left to right in an \emph{incremental} manner ($L_{K}$, $H_{K-1}$, $H_{K-2}$, ...), and every time has the loss defined for the corresponding Gaussian pyramid level, e.g. $loss(A_K, \widetilde{A}_{K})$, $loss(A_{K-1}, \widetilde{A}_{K-1})$, $loss(A_{0}, ...\widetilde{A}_{0})$ for the albedo network. This incremental training constrains the network to output a near-perfect Gaussian pyramid, and that the sub-network $H_{i}, i=K-1,...0$ outputs the expected Laplacian detail layer. Figure~\ref{fig:illustration} shows intermediate outputs of the network trained in this scheme for illustration. Except we state otherwise, the quantitative results are obtained using the simultaneous training scheme.

\subsection{Self-Augmented Training}
In this section, we describe a data augmentation strategy for incorporating unlabeled images to self-augment our network training process. We draw the inspiration from the work of \textit{breeder learning}~\cite{nair2008analysis}. The idea is to employ a forward generative model to generate new training pairs for a model by perturbing parameters produced by the model to be augmented. This mechanism bears the spirit of Boostrap to some extent  and turns out to be quite effective. For example, Li et al.~\cite{li2017modeling} recently applied this strategy in an appearance modeling network by generating training images from model's predicted reflectance of unlabeled images.
%leverage the appearance information embedded in unlabeled images to
%Recently, \cite{li2017modeling} propose to leverage the appearance information embedded in unlabeled images of spatially varying materials to self-augment the training process in order to reduce the amount of required labeled training data.

We start with a preliminary network trained with a moderately sized dataset that has ground-truth albedo and shading. We then apply the network to a set of new images and obtain the estimated albedo $\tilde{A}$ and shading $\tilde{S}$. With a straightforward synthesis procedure, we can generate a new image from the estimations. Note that by our loss definitions, $\tilde{A}$ and $\tilde{S}$ are not hard constrained to exactly match the input image (as in \cite{lettry2016darn}), so the new synthesized images will deviate from the original ones.

\begin{figure}
  \centering
  \begin{overpic}[width=1.0\linewidth,clip]{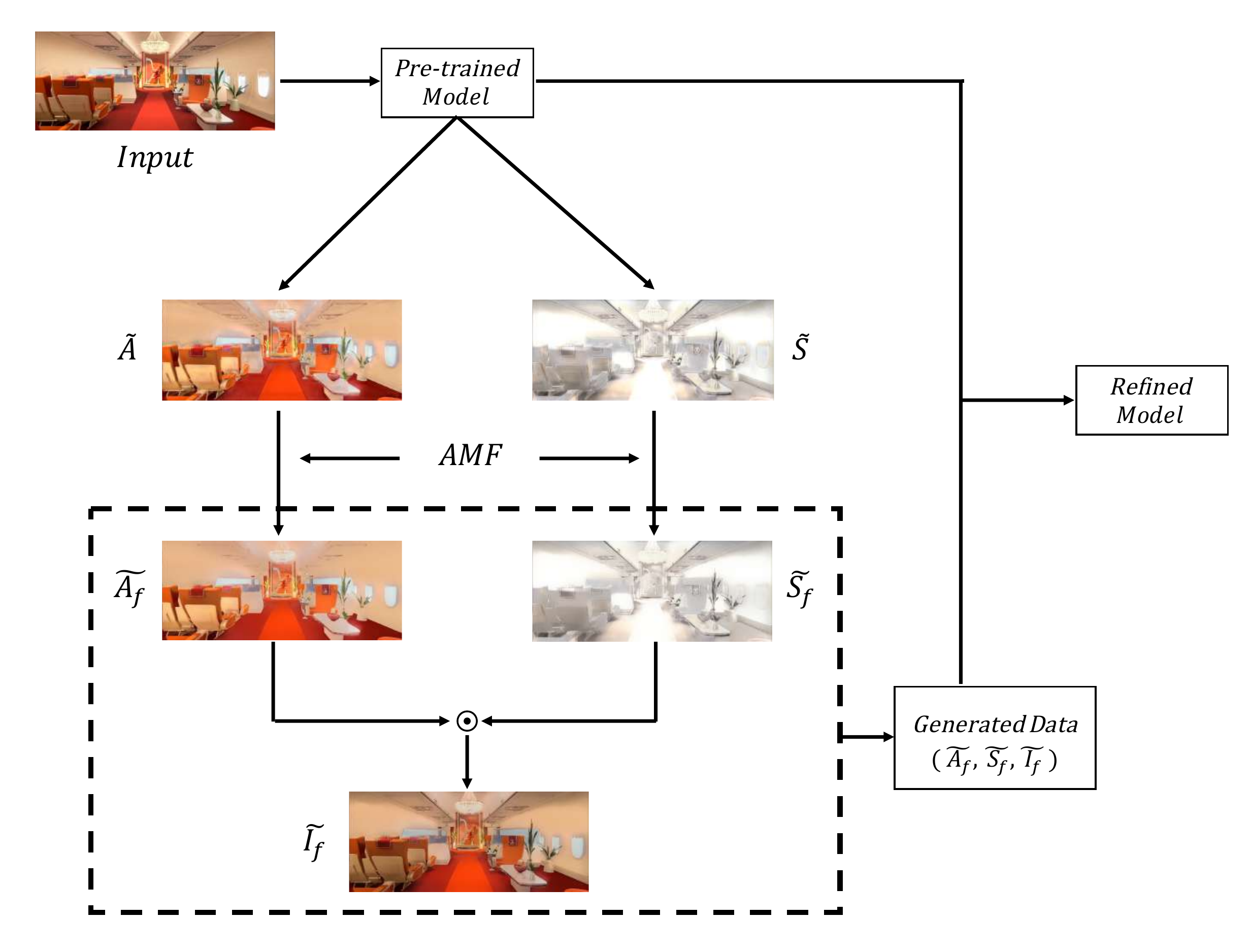}\end{overpic}\\
  \caption {Our data augmentation process uses a preliminarily train model to produce estimations for unlabeled data, and use the estimation result to \emph{synthesize} new data for self-augmented training.}\label{fig:refine}\vspace{-5mm}
\end{figure}

To introduce further perturbation in the augmented dataset, we additionally apply an \textit{Adaptive Manifold Filtering} (AMF, ~\cite{Gastal2012Adaptive}) operation to $\tilde{A}$ and $\tilde{S}$ and use the filtered results to synthesize new data (see Figure~\ref{fig:refine}). The AMF filtering operator suppresses noise or unwanted details in  $\tilde{A}$ and $\tilde{S}$ that may come from the input images or produced by the premature network, and serves to ``regularize'' the manifold of the new synthesized images and their ground-truth label space so that the network is not misled to overfit capricious details in the self-augmented training process.

%\input{method_sub}
%%%%%%%%%%%%%%%%%%%%%%%%%%%%%%%%%%%%%%%%%%%%%%%%%%%%%%
%%%%%%%%%%%%%%%%%%%%%%%%%%%%%%%%%%%%%%%%%%%%%%%%%%%%%%
%%%%%%%%%%%%%%%%%%%%%%%%%%%%%%%%%%%%%%%%%%%%%%%%%%%%%%
%%\input{experiment}

\section{Evaluation}
%We conducted evaluation of our model two major benchmark datasets: the MPI-Sintel dataset and the MIT-Intrinsic Images dataset. The results are in Table~\ref{tab:sintel_image_split}-\ref{tab:mit_scenes} and Figure~\ref{fig:comparison_MIT}-\ref{fig:comparison}. The results show that our model significantly outperform the most recent neural network based models and traditional methods. Results also show, with extensive self-comparisons, the effectiveness of individual modules of our model, including the pyramid structure and data augmentation.
In this section we describe evaluation of the model on the MPI-Sintel dataset and the MIT Intrinsic Images dataset and show results in Table~\ref{tab:sintel_image_split}-\ref{tab:mit_scenes} and Figure~\ref{fig:comparison_MIT}-\ref{fig:comparison}.

%This section is organized as follows. We first describes our implementation details and experiment setup. Then, we analyze and evaluate the proposed network in various aspects.

\subsection{Experiment Setup}

\noindent
{\bf DataSet and Metrics}
The MPI-Sintel dataset\cite{butler2012naturalistic} is composed of 18 scene level computer generated image sequences, 17 of which contain 50 images of the scene and one contains 40 images. We follow~\cite{narihira2015direct,lettry2016darn} and use the $ResynthSintel$ version in our experiment because the data satisfies the $\mathcal{A} \times \mathcal{S} = \mathcal{I}$ constraint. Two types of train/test split (\textit{scene split} and \textit{image split}) are used for head-to-head comparison with previous work. The \textit{scene split} splits the dataset at scene level which takes half of the scenes for training and the rest scenes for testing. The \textit{image split} randomly pick half of the images for training/testing without considering their scene category. The original version of the MIT Intrinsic dataset~\cite{Grosse:2009} has 20 object-level images taking in a laboratory environment setup, each with 11 different lighting conditions. We use the same strategy of \cite{BarronTPAMI2015} to split the data for direct comparison.

\noindent Evaluations are based on the following metrics:\vspace{-2mm}
\begin{description}
\item[si-MSE] scale-invariant mean squared error (si-MSE) defines the pixel-wise MSE up to a free scaling factor (see \cite{BarronTPAMI2015}).
\item[si-LMSE] scale invariant local mean square error (si-LMSE) measures the averaged si-MSE on local window patches as the window slides over the image with a stride. The window size is usually set to 10\% of the image size along the larger dimension and stride is half of the window size:
%we usually set the local window size $k$ as 10\% of the image along its larger dimension. $N_{\mathcal{W}}$ denotes the number of local windows and the definition is:
%\begin{equation}
\[\begin{aligned}
& \text{si-LMSE}(\mathcal{C}_{gt},\widetilde{\mathcal{C}}) =\frac{1}{N_{\mathcal{W}}} \sum_{\omega\in\mathcal{W}}\text{si-MSE}(\mathcal{C}_{gt}^{\omega},\widetilde{\mathcal{C}}^{\omega})
\end{aligned}\]
\item[LMSE] The LMSE measure is the ``normalized'' si-LMSE measure on albedo and shading together. We use this metric on the MIT Intrinsic Images dataset. Local window size for si-LMSE is set to 20 (as in \cite{grosse09intrinsic}):
%Here the metric {\bf LMSE} is based on ~\cite{grosse09intrinsic} and the local window size is set to 20. The formulation is detailed as:
\[
\begin{aligned}
& \text{LMSE} = \frac{1}{2}\frac{\text{si-LMSE}(\mathcal{S}_{gt},\widetilde{\mathcal{S}})}{\text{si-LMSE}(\mathcal{S}_{gt},0)} + \frac{1}{2}\frac{\text{si-LMSE}(\mathcal{A}_{gt},\widetilde{\mathcal{A}})}{\text{si-LMSE}(\mathcal{A}_{gt},0)}
\end{aligned}
\]
\item[DSSIM] The structural similarity is quantized by \textit{dissimilarity structural similarity index measure} as $\frac{1-\text{SSIM}}{2}$ (see \cite{zhou:ssim04} for SSIM definition).
\end{description}

\noindent
{\bf Implementation Details}
We implemented our model in the PyTorch framework with mini-batch size 8.
In training, we get the input image by randomly cropping patches of size 256 $\times$ 256 after scaling by a random factor in [0.8,1.2] and using random horizontal flipping with probability 0.5. We empirically construct 4 levels of pyramids and initialize all the weights with the strategy of~\cite{he2015delving}. Besides, we adopt the Adam~\cite{kingma2014adam} optimization method with a learning rate starting at $10^{-4}$ and decreasing to $10^{-6}$. We use 2x the size of the training data as the size of the augmentation data in both experiments.

\begin{table*}[htb]
\centering
\begin{tabular}{cccccccccc}
\hline
\multirow{2}{*}{Sintel \textit{image split}}&
\multicolumn{3}{c}{si-MSE}&\multicolumn{3}{c}{si-LMSE}&\multicolumn{3}{c}{DSSIM}\cr\cline{2-10}
&A&S&avg&A&S&avg&A&S&avg\cr\hline\hline
Baseline: Constant Shading&5.31&4.88&5.10&3.26&2.84&3.05&21.40&20.60&21.00\cr%\hline
Baseline: Constant Albedo&3.69&3.78&3.74&2.40&3.03&2.72&22.80&18.70&20.75\cr%\hline
Color Retinex~\cite{grosse09intrinsic} &6.06&7.27&6.67&3.66&4.19&3.93&22.70&24.00&23.35\cr%\hline
Lee et al.~\cite{Lee:2012:EII} &4.63&5.07&4.85&2.24&1.92&2.08&19.90&17.70&18.80\cr%\hline
Barron \& Malik~\cite{BarronTPAMI2015} &4.20&4.36&4.28&2.98&2.64&2.81&21.00&20.60&20.80\cr%\hline
Chen and Koltun~\cite{Chen:iccv13}&3.07&2.77&2.92&1.85&1.90&1.88&19.60&16.50&18.05\cr%\hline
Direct Intrinsic~\cite{DirectIntrinsic:2015}&1.00&0.92&0.96&0.83&0.85&0.84&20.14&15.05&17.60\cr%\hline
DARN~\cite{DARN16}&1.24&1.28&1.26&0.69&0.70&0.70&12.63&12.13&12.38\cr%\hline
Kim et al.~\cite{DBLP:KimPSL16}&0.7&0.9&0.7&0.6&0.7&0.7&9.2&10.1&9.7\cr%\hline
Fan et al.~\cite{DBLP:FanWHC17}&0.67&0.60&0.63&0.41&0.42&0.41&10.50&7.83&9.16\cr\hline\hline
Ours Sequential &0.83&0.74&0.79&0.58&0.54&0.56&7.61&7.91&7.76\cr %2017-10-20-11-41
Ours Hierarchical & 0.81 & 0.78 & 0.79 & 0.58 & 0.58 & 0.58 & 8.18 & 7.16 & 7.62 \cr
Ours w/o Pyramid&0.92&1.37&1.15&0.65&1.15&0.90&8.44&10.96&9.70\cr%\hline %2017-10-16-01-02
Ours w/ MSE loss&0.72&0.62&0.67&0.62&0.46&0.50&7.98&6.37&7.18\cr %2017-10-16-00-33
%\color{red} Ours w/o Shading TV loss&0.71&0.58&0.64&0.48&0.39&0.44&6.81&6.09&6.45\cr
Ours w/ `FPN' input &0.73&0.60&0.67&0.49&0.43&0.46&6.84&6.76&6.80\cr\hline
Ours Final*&0.66&0.60&0.63&0.44&0.42&0.43&6.56&6.37&6.47\cr
Ours Final+DA&\bf{0.61}&\bf{0.57}&\bf{0.59}&\bf{0.41}&\bf{0.39}&\bf{0.40}&\bf{5.86}&\bf{5.97}&\bf{5.92}\cr\hline% 2017-10-10_12-10
\end{tabular}
\caption{Quantitative Evaluation ($\times$100) on the MPI-Sintel \textit{image split}}\label{tab:sintel_image_split}
\end{table*}

\begin{table*}[htb]
\centering
\begin{tabular}{cccccccccc}
\hline
\multirow{2}{*}{Sintel \textit{scene split}}&
\multicolumn{3}{c}{si-MSE}&\multicolumn{3}{c}{si-LMSE}&\multicolumn{3}{c}{DSSIM}\cr\cline{2-10}
&A&S&avg&A&S&avg&A&S&avg\cr\hline\hline
Direct Intrinsic~\cite{DirectIntrinsic:2015}&2.01&2.24&2.13&1.31&1.48&1.39&20.73&15.94&18.33\cr%\hline
DARN~\cite{DARN16}&1.77&1.84&1.81&0.98&0.95&0.97&14.21&14.05&14.13\cr%\hline
Fan et al.~\cite{DBLP:FanWHC17}&1.81&1.75&1.78&1.22&1.18&1.20&16.74&13.82&15.28\cr\hline\hline
Ours Sequential & 1.61 & 1.56 & 1.58 & 1.05 & 1.11 & 1.08 & 10.24 & 11.90 & 11.07 \cr %2017-10-21_18-02
Ours Hierarchical&1.59&1.51&1.55&0.98&1.01&0.99&8.70&9.55&9.13\cr
Ours w/o Pyramid&1.82&2.01&1.92&1.01&1.39&1.20&14.43&14.27&14.35\cr%\hline
Ours w/ MSE loss &1.47&1.44&1.46&0.92&0.95&0.93&9.48&10.97&10.23\cr
%\color{red} Ours w/o Shading TV loss&1.37&1.35&1.36&0.89&0.92&0.91&8.58&9.28&8.93\cr
Ours w/ `FPN' input  &1.46&1.40&1.43&0.96&0.97&0.97&8.50&9.30&8.90\cr\hline
Our Final* &1.38&1.38&1.38&0.92&0.93&0.92&8.46&9.26&8.86\cr
Our Final+DA &\bf{1.33}&\bf{1.36}&\bf{1.35}&\bf{0.82}&\bf{0.89}&\bf{0.85}&\bf{7.70}&\bf{8.66}&\bf{8.18}\cr\hline %%  2017-09-12-21-12 (back->2017-09-14)
\end{tabular}
\caption{Quantitative Evaluation ($\times$100) on the MPI-Sintel \textit{scene split}}\label{tab:sintel_scene_split}
\end{table*}

\iffalse
\begin{figure}
\centering
\includegraphics[width=\linewidth]{hierarchical_optimization.png}
\caption{hierarchical optimization in each level}\label{fig:hierarchical}
\end{figure}
\fi

\subsection{Evaluation on MPI-Sintel Dataset}
\begin{figure*}
\centering
\vspace{-8mm}
\includegraphics[width=0.82\textwidth,clip,trim=0 50 0 25]{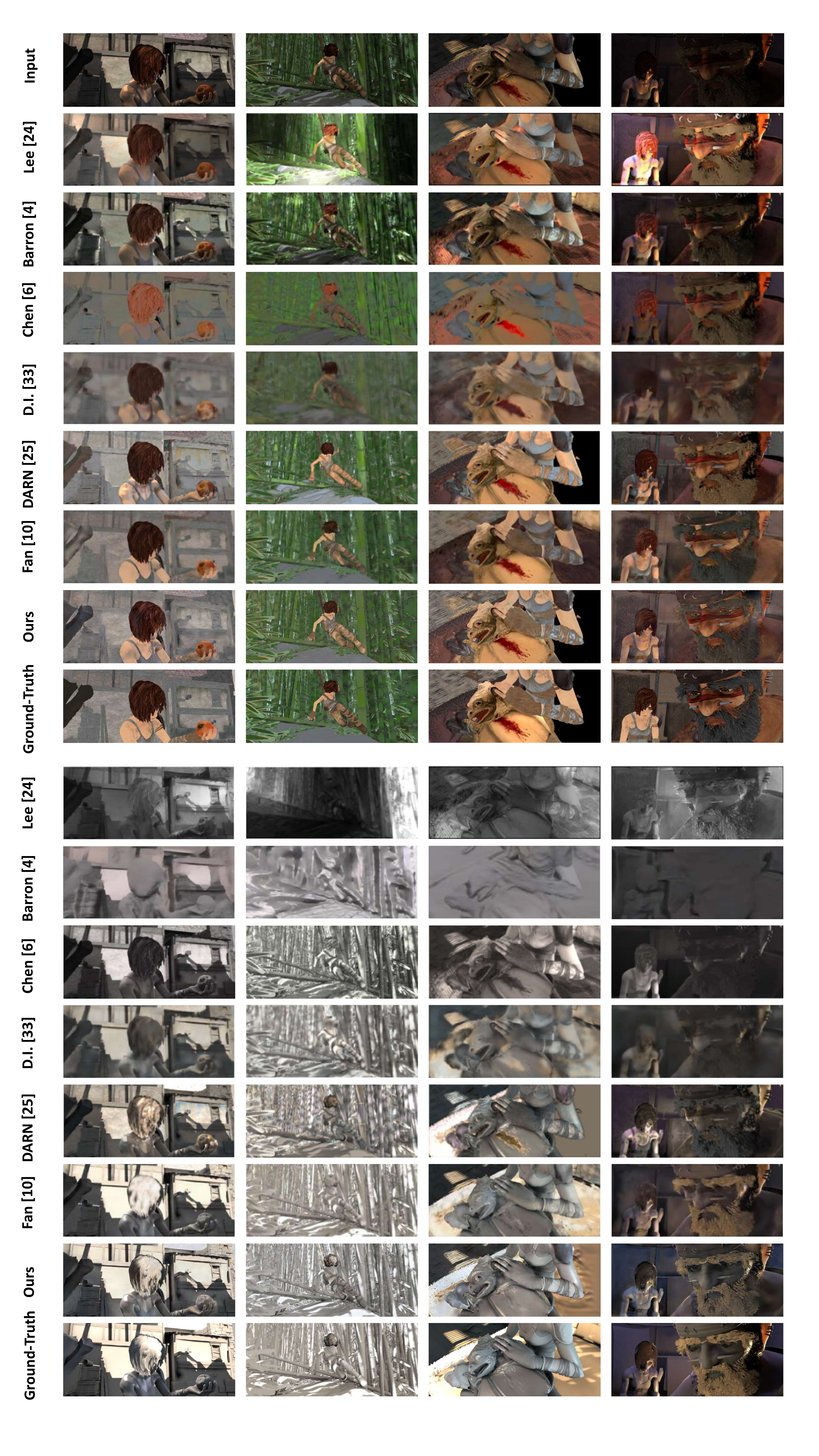}
\caption{Qualitative results on four examples of the MPI-Sintel benchmark dataset and comparison to previous methods (results are excerpted from paper with limited resolution). Notice our decomposition results exhibit good edge preserving property and are visually close to the ground truth. }\label{fig:comparison}
\end{figure*}

The evaluation results on the MPI-Sintel dataset are in Table~\ref{tab:sintel_image_split}-\ref{tab:sintel_scene_split} and Figure~\ref{fig:comparison}. Again, our model produces favorable results over previous methods, especially in the \emph{scene split} where the network is less prone to ``overfit'' for the test data.

\noindent
{\bf Comparison with Previous Work:}
%We first compare our results with existing methods on the scale invariant metrics mentioned before~(Table~\ref{tab:sintel_image_split},Table~\ref{tab:sintel_scene_split},Figure~\ref{fig:comparison}). The numbers in the two tables clearly show that deep learning methods~\cite{DirectIntrinsic:2015,DARN16,DBLP:KimPSL16,DBLP:FanWHC17} greatly improve the results compared to traditional methods~\cite{grosse09intrinsic,Lee:2012:EII,BarronTPAMI2015,Chen:iccv13}.
We first compare our model with a series of previous methods, including the two naive baselines \emph{Constant Shading} and \emph{Constant Albedo}, a few of the traditional methods (\cite{grosse09intrinsic,Lee:2012:EII,Chen:iccv13,BarronTPAMI2015}), and the recent up-to-date neural network based models (\cite{DirectIntrinsic:2015,DARN16,DBLP:KimPSL16,DBLP:FanWHC17}). The result shows our model with/without data augmentation both yield new state-of-the-art performance across all the three metrics. %The advantage is more convincing on the structure-based DSSIM metric.
We do want to point out the quantitative result of all methods (including ours) on the Sintel \emph{image split} might be misleading to some extent. This is because the image sequences of the same scene category in the Sintel dataset are very similar to each other, so by splitting all the data at image level (images of the same scene type may appear in both train and test sets), an over-fit network on the training set will still appear to ``perform'' well on the test set. But the \emph{scene split} dataset will not have this problem. An interesting result in the Tables is that the margin of our results to previous results is larger in the \emph{scene split} (Table~\ref{tab:sintel_scene_split}) than the \emph{image split} (Table~\ref{tab:sintel_image_split}).  In the Tables, even though we hold a fairly moderate margin on the \emph{image split}, the margin we hold on the \emph{scene split} is up to $25\%$ in si-MSE and $43\%$ in DSSIM, showing that our network can generalize significantly better for this more challenging data split.%better than the previous methods.
%A reasonable explanation to this is that the training and testing sets of the image split are close, and an over-fit network always performs well. While performing on the scene split is more challenging as it requires more generalization capacity. As shown in the Tables, methods like~\cite{DBLP:FanWHC17} are not convincing enough for exhibiting good performance in Table~\ref{tab:sintel_image_split} while showing poor quantitative results in Table~\ref{tab:sintel_scene_split}. On the contrary, our results show excellent quantitative results in both two types of data split ({\bf Ours Final+DA}) and visual effect (see Figure~\ref{fig:comparison}) as well.

\noindent
{\bf From Sequential to Parallel  Architecture:}
An important network architecture reformation we described in section~\ref{sec:reformation} is from the sequential
structure to the multi-branch parallel structure (Figure~\ref{fig:reformation}-(a) to (c)). This reformation flattens a deeply stacked network into a set of parallel channels, therefore alleviates the issues of gradient back-propagation. The row ({\bf Ours Sequential}) displays the result by the sequential architecture (a) in Figure~\ref{fig:reformation}.  It shows this architecture produces comparable performance against previous works, but suboptimal to our final model, especially in the DSSIM metric (7.76 and 11.07 down to 6.47 and 8.86). %As we discussed in section~\ref{sec:reformation}, an important conversion of Laplacian Pyramid is to change architecture from sequential ({\bf Ours Sequential}) to parallel ({\bf Ours Final*}) as depicted in Figure~\ref{fig:reformation}. From Table~\ref{tab:sintel_image_split} and Table~\ref{tab:sintel_scene_split}, we found that the parallel architecture is preferable. We offer a possible explanation under the circumstances: each sub-network in sequential architecture gets input from the output of the previously sub-network and thus cause relatively poor convergence due to the strong constraint in each level. However, we still can not jump to conclusions due to the specific application and more research is needed to explore this subject.

\noindent
{\bf Hierarchical Optimization \textit{vs} Joint Optimization:}
Another architectural optimization in our work is removing the constraint (loss) at each Laplacian pyramid level (Figure~\ref{fig:reformation}-(c)), and simultaneously train all the network channels with a single loss constraint (Figure~\ref{fig:reformation}-(d)). We call the optimization scheme in the latter case \emph{joint optimization}, and that of the former \emph{hierarchical optimization}. A figure is included in the supplemental material explaining more details of the hierarchical optimization. In Table~\ref{tab:sintel_image_split}-\ref{tab:sintel_scene_split}, it shows a $10\%-15\%$ improvement by the joint optimization scheme across all metrics.
%A natural intuition to optimize pyramid architecture is to minimize the difference between Gaussian images at each level (Figure~\ref{fig:reformation} (c)). As we can see in Figure~\ref{fig:hierarchical}, the accumulative error is propagated from the current level to the next. In the procedure of hierarchical optimization, the output of each sub-network is restrained to corresponding Laplacian Pyramid component. We further extend this strategy to a more relaxed situation that we only pay attention to the final reconstruction image named jointly optimization. The experiments in Table~\ref{tab:sintel_image_split} and Table~\ref{tab:sintel_scene_split} reveal that our jointly optimization ({\bf Ours Final*}) method is better in this task in contrast to hierarchical optimization ({\bf Ours Hierarchical}).

\noindent
{\bf Self-Comparison on other Factors:}
We also have a set of controlled self-comparison with respect to other factors, including the \emph{pyramid structure}, \emph{loss function}, \emph{alternating network input}, and \emph{data augmentation}. \\%Qualitative results of self-comparisons are included in the supplemental material.\\
%The model we proposed shows good results both quantitatively and perceptually. While it is desirable to prove the validity of the architecture in some key aspects. For more details, please refer to supplementary materials.\\
\noindent
{\bf \textit{Pyramid structure}}
The row ({\bf Ours w/o Pyramid}) displays result using a single-channel network, i.e. we use a single residual block to produce output from input directly without having the multi-band decomposition structure. The results in Table~\ref{tab:sintel_image_split} and Table~\ref{tab:sintel_scene_split} show that our counterpart model with the pyramid structure improves over than \emph{30\%} compared to controlled setting by turning this feature off. Note the network complexity grows sub-linearly up to a constant factor as the number of pyramid layer increases.\\ %convolutional network and exhibits good property in keeping the distribution of final prediction map.\\
%The main framework of our model is based on pyramids. We just make the change to utilize the architecture of one pyramidal block ({\bf Ours w/o Pyramids}) for comparison. The contrastive results in Table~\ref{tab:sintel_image_split} and Table~\ref{tab:sintel_scene_split} show that our pyramids architecture improves more than 30\% compared to naive convolutional network and exhibits good property in keeping the distribution of final prediction map.\\
\noindent
{\bf \textit{Loss function:}}
The row ({\bf Ours w/ MSE loss}) displays result by replacing our loss function with the classical MSE loss. It turns out the quantitative error with the MSE loss does not degrade by a large factor in the scale-invariant MSE metrics. However, qualitative results in supplemental material do reveal the MSE loss produces results with blurry edges. The structure-based metric (DSSIM) also shows a clearer margin (from 10.23 to 8.86 in the scene split) between the MSE loss and our loss.\\%The numbers of quantitative results ({\bf Ours w/o Elaborate loss}) are close to our model ({\bf Ours Final*}) in Table~\ref{tab:sintel_image_split} and Table~\ref{tab:sintel_scene_split}, while Figures in supplementary materials indicated that our loss function make a great progress in visual quality.\\
%It is plausible that we can compute loss by employing mean square error(MSE) but it is true that this often blurred output image. In this work, we propose an elaborated loss function which not only in terms of properties of an intrinsic image but also considering perceptual quality. The numbers of quantitative results ({\bf Ours w/o Elaborate loss}) are close to our model ({\bf Ours Final*}) in Table~\ref{tab:sintel_image_split} and Table~\ref{tab:sintel_scene_split}, while Figures in supplementary materials indicated that our loss function make a great progress in visual quality.\\
\noindent
{\bf \textit{CNN features as input}}
We further investigate the affect of having Gaussian pyramid image components as input of our network in this task, as most existing multi-scale deep networks (e.g. \cite{lin2016feature,pinheiro2015learning,ghiasi2016laplacian,lai2017deep}) use multi-scale features produced by a CNN network. The row ({\bf Ours w/ `FPN' input}) shows the result that takes CNN features as input following exactly the FPN network~\cite{lin2016feature}. The comparison shows our final model holds a slight but unclear advantage, meaning that the high level features of a CNN still well preserve much of the necessary semantic information for our \emph{pixel-to-pixel} transformation network.\\
%We further investigate the problem of the output at each pyramidal level. To our knowledge, most recently pyramid based methods used feature maps as output in network, e.g. {FPN}. In our architecture, we enforce the sub-network to produce corresponding Laplacian images. For fair comparison, we control variable to change the output to the last feature maps ({\bf Ours FPN}) in the network. Notice that {\bf Ours FPN} is different from naive FPN due to our parallel architecture. The experiments in Table~\ref{tab:sintel_image_split} and Table~\ref{tab:sintel_scene_split} demonstrate that our scheme is more effective.\\
\noindent
{\bf \textit{Data augmentation}}
The last row in Table~\ref{tab:sintel_image_split}-\ref{tab:sintel_scene_split} shows the effect of our data augmentation. We obtain a set of cartoon clips crawled from the Web that share similar property with the MPI dataset (see an example in Figure~\ref{fig:refine}). The size of the augmentation data is set to 2 times of the labeled training data. Further increasing the augmentation data size did not produce important improvement in our experiment.
%Noted that our data augmentation is different from usual data preprocessing like rotation, translation. Instead, here we refer to a more general case that we breeder more training pairs from unlabelled data. In our experiments, we obtain a set of cartoon clips from the internet and decompose them into unlabelled input images. The experiments show that this kind of method ({\bf Ours Final+DA}) improves the final results. In addition, we test the settings with different size of data, and find the results are not getting better with more unlabelled data. We speculate that massive unlabelled data were dominant enough to shift the distribution of naive MPI data set. We empirically set unlabelled data to 2 times or so than original training data.

\subsection{Evaluation on MIT Intrinsic Images Dataset}
\begin{figure}
\centering
%\begin{overpic}[width=\textwidth,height=1.8\textwidth]{Comparison_new.png}\end{overpic}\\
\includegraphics[width=\linewidth,clip,trim=0 180 1200 0]{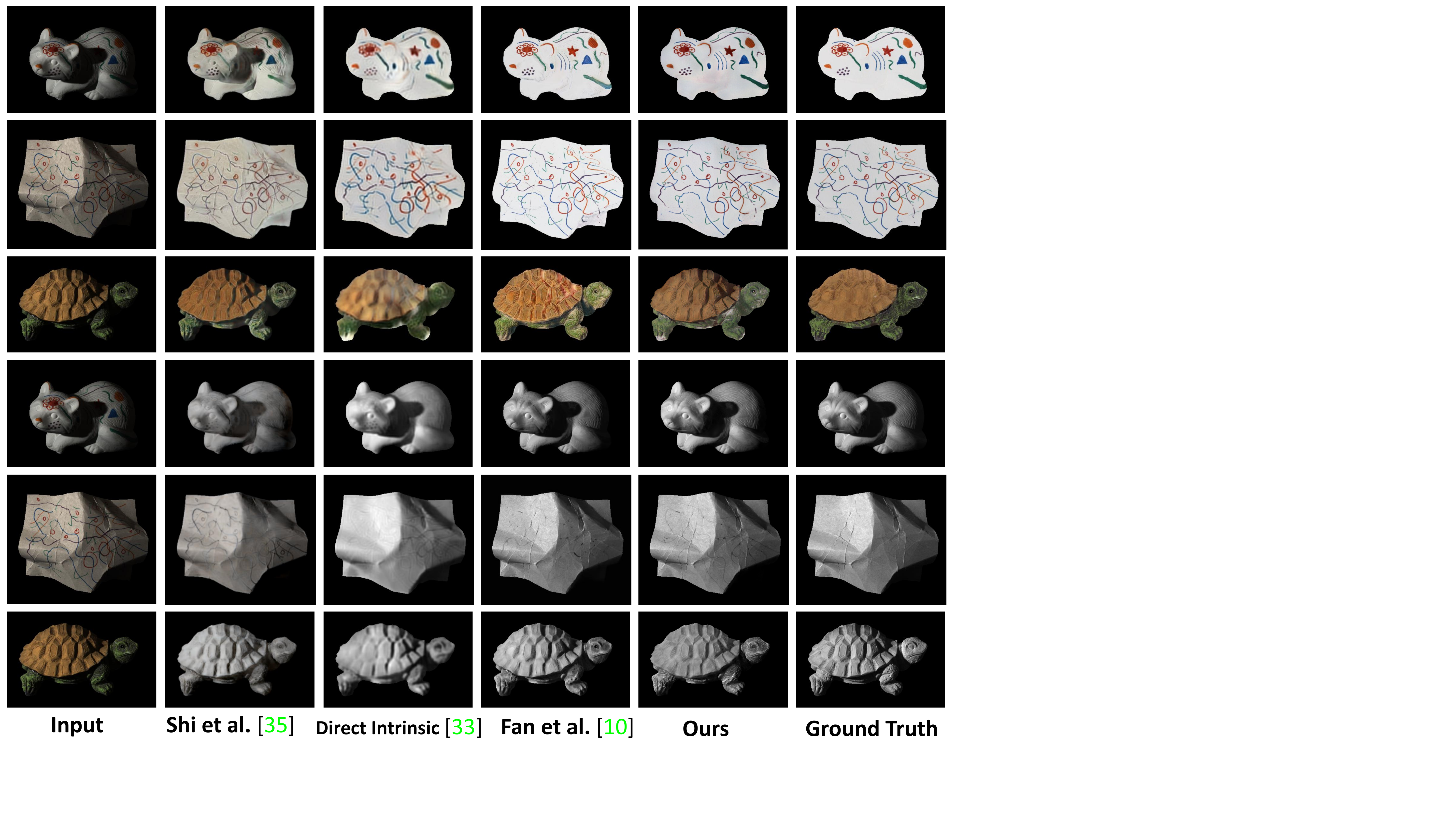}
\caption{Qualitative results on the MIT Intrinsic dataset examples. Top three rows are albedo; the bottom three rows are shading.}\label{fig:comparison_MIT}\vspace{0mm}
\end{figure}

We also evaluated the performance of our model against a set of previous methods
on the {MIT Intrinsic Images dataset}.  The results are shown in Table~\ref{tab:mit_scenes} and Figure~\ref{fig:comparison_MIT}. %Note that our model (without data augmentation) is the first neural network based model that achieves the performance reported in Barron and Malik~\cite{BarronTPAMI2015} in averaged si-MSE.
In this set of experiments, we conducted data augmentation in two different setups: {\bf Ours + DA} and {\bf Ours + $\text{DA}^{+}$}. The difference is in the \emph{data} that we take for the augmentation. {\bf Ours + DA} is by the ordinary setting where the augmenting data is searched from the web by a set of similar object category names the dataset provides. In {\bf Ours + $\text{DA}^{+}$}, instead, we generate the augmenting dataset from the same set of objects (depth and reflectance) of the MIT Intrinsic Images dataset under new illumination conditions (spherical harmonic illuminations from \cite{Barron:2012B} and the rendering method by \cite{Ramamoorthi:2001:SH}). This creates a dataset that highly resembles the original dataset and is practically impossible to acquire in real case. In other words, it sets a ceiling for the quality of augmentation data. The results in Table~\ref{tab:mit_scenes} shows that both augmentation setups are effective, and the latter one gives clue to the limit we can get from the data augmentation scheme we introduced for this task.
\begin{table}[t]
\resizebox{\linewidth}{!}{
\begin{tabular}{ccccc}
\hline
\multirow{2}{*}{\bf{Mit Intrinsic Data}}&
\multicolumn{3}{c}{si-MSE}&\multicolumn{1}{c}{LMSE}\cr\cline{2-5}
&Albedo&Shading&Average&Total\cr\hline\hline
Zhou \etal ~\cite{zhou2015learning}&0.0252&0.0229&0.0240&0.0319\cr%\hline
Barron \etal~\cite{BarronTPAMI2015}&0.0064&0.0098&0.0081&0.0125\cr%\hline
Shi \etal~\cite{shi2016learning}&0.0216&0.0135&0.0175&0.0271\cr%\hline
Direct Intrinsic \etal~\cite{DirectIntrinsic:2015}&0.0207&0.0124&0.0165&0.0239\cr%\hline
Fan \etal~\cite{DBLP:FanWHC17}&0.0127&0.0085&0.0106&0.0200\cr\hline
Ours*&0.0089&0.0073&0.0081&0.0141\cr%2017-11-13-17-07
Ours + DA &0.0085&0.0064&0.0075&0.0133\cr\hline %2017-11-14-00-42
Ours + $\text{DA}^{+}$ &0.0074&0.0061&0.0068&0.0121\cr\hline %2017-11-13-22-51
\end{tabular}
}
\caption{Evaluation on the {\bf MIT Intrinsic Images dataset}. }\label{tab:mit_scenes}%Noted that \textit{LMSE} here is computed based on~\cite{grosse09intrinsic} for fair comparison}
\end{table}

%Note that Barron\etal~\cite{BarronTPAMI2015} uses additional priors to boost their performance and our results outperform best to other deep learning methods.
%The experiment demonstrates that our model is flexible and robust enough in intrinsic image decomposition problem (See figure~\ref{fig:comparison} for visual results).

%Besides, we adopt two types of data augmentation strategies: a) we scratch new categories resemble to MIT data to refine our model ({\bf Ours + DA}). b) each training object in the MIT dataset was randomly assigned an illumination (such that training illuminations were assigned to training objects, etc), and was re-rendered under its new illumination to get our unlabelled augmented data ({\bf Ours + $\text{DA}^{+}$}). The results show that both two strategies improve the result and the second way under different illumination are highlighted. We believe this is because the re-rendered MIT data shares similar distribution with our naive training data and determines the upper bound of the convergence. While the data we scratched from web is uncontrollable due to its complicated illumination.

\section{Conclusion}
%This paper presents a novel framework for intrinsic image decomposition which learns the image-to-image transformation function in successive frequency bands in parallel by expanding the output images (albedo and shading) into Laplacian pyramid components. To alleviate the drawback of insufficient data, we use the mechanism inspired by \textit{breeder learning} with adaptation to our task. On both MPI Sintel and MIT dataset, the proposed method not only outperforms the state of the art by a remarkable margin, but also yields promising qualitative results. These experimental results clearly demonstrate the significance of architecture and strategy.
%conclusion.
%We introduce a new network structure for decomposing an image into its intrinsic albedo and shading. We treat this as an image-to-image transformation problem and explore the scale space of the input and output. By expanding the output images (albedo and shading) into their Laplacian pyramid components, we develop a multi-channel network structure that learns the image-to-image transformation function in successive frequency bands in parallel, within each channel is a fully convolutional neural network with skip connections. This network structure is general and extensible, and has demonstrated excellent performance on the intrinsic image decomposition problem.  We evaluate the network on two benchmark datasets: the MPI-Sintel dataset and the MIT Intrinsic Images dataset. Both quantitative and qualitative results show our model delivers a clear progression over state-of-the-art.
We have introduced a Laplacian pyramid inspired neural network architecture for intrinsic images decomposition. The network models the problem as image-to-image transformation and expands the input and output in their scale space. We have conducted experiments on the MPI Sintel and MIT dataset and produced state-of-the-art quantitative results and good qualitative results. For future work, we expect the proposed network architecture to be tested and refined on other image-to-image transformation problems, e.g., pixel labeling or depth regression.

\vspace{3mm}
\noindent{\bf Acknowledgment}
We thank all the anonymous reviewers. This work is supported in part by National Key R$\&$D Program of China (No. 2017YFB1002703), by NSFC (No. 61602406), by ZJNSF (No. Q15F020006), and by a special fund from the Alibaba - ZJU Joint Institute of Frontier Technologies.

{\small
\bibliographystyle{ieee}
\bibliography{zcl}
}

\end{document}